\documentclass[10pt,journal,compsoc]{IEEEtran}

\usepackage{times}
\usepackage{epsfig}
\usepackage{graphicx}
\usepackage{amsmath}
\usepackage{amssymb}
\usepackage{booktabs}
\usepackage{etex}
\usepackage{footnote}
\usepackage{tablefootnote}

\usepackage{algorithmic}

\usepackage{graphicx}
\usepackage{latexsym}
\usepackage{amsmath}
\usepackage{amssymb}
\usepackage{multirow}

\usepackage{color}
\usepackage[dvipsnames]{xcolor}

\usepackage[ruled,linesnumbered]{algorithm2e}
\usepackage{epstopdf}
\usepackage{array}
\newcolumntype{C}[1]{>{\centering\arraybackslash}m{#1}}

\newcommand{\gr}{{\rm grad}}

\SetKwInput{KwInput}{Input}
\SetKwInput{KwOutput}{Output}

\usepackage{url}

\usepackage{amssymb}
\usepackage{amsfonts}
\usepackage{amsthm}
\usepackage{MnSymbol}

\usepackage[cmtip,all]{xy}
\usepackage[all,cmtip]{xy}

\newcommand{\openM}{\mathring{\mathcal{M}}}

\newcommand{\hatM}{\hat{\mathcal{M}}}
\newcommand{\setM}{\mathcal{M}}

\newcommand{\setL}{\mathcal{L}}

\newtheorem{theorem}{Theorem}[section]

\newtheorem{prop}[theorem]{Proposition}

\theoremstyle{definition}
\newtheorem{definition}[theorem]{Definition}
\theoremstyle{definition}

\theoremstyle{remark}

\theoremstyle{remark}

\theoremstyle{remark}

\newcommand*{\QEDbs}{\hfill\ensuremath{\blacksquare}}%
\usepackage{animate}

\usepackage{graphicx}

\usepackage[caption=false]{subfig}
\usepackage[T1]{fontenc}

\newcounter{mySentence}                               
\newcounter{myTempSentence}                           
\renewcommand*{\themySentence}{\textbf{(\arabic{mySentence})}}               
\newenvironment{sentence}[1][]{
	\begin{list}{\themySentence}{\leftmargin=-1cm \itemindent=-1cm}{\refstepcounter{mySentence}}\item 
		\ifnum\pdfstrcmp{#1}{}=0\else\label{#1}\fi                     
	}{\end{list}}

\usepackage[pagebackref=true,breaklinks=true,letterpaper=true,colorlinks,bookmarks=false]{hyperref}

  \graphicspath{{images/}}
  
  \DeclareGraphicsExtensions{.pdf,.jpeg,.png,.eps}

\begin{document}



\title{Optimization on Submanifolds of \\ Convolution Kernels in CNNs}

\author{Mete~Ozay and Takayuki~Okatani\\
Graduate School of Information Sciences, Tohoku University, Sendai, Miyagi, Japan.
{\tt\small \{mozay,okatani\}@vision.is.tohoku.ac.jp} }

\maketitle

\begin{abstract}
   Kernel normalization methods have been employed to improve robustness of optimization methods to reparametrization of convolution kernels, covariate shift, and to accelerate training of Convolutional Neural Networks (CNNs). However, our understanding of theoretical properties of these methods has lagged behind their success in applications. We develop a geometric framework to elucidate underlying mechanisms of a diverse range of kernel normalization methods. Our framework enables us to expound and identify geometry of space of normalized kernels. We analyze and delineate how state-of-the-art kernel normalization methods affect the geometry of search spaces of the stochastic gradient descent (SGD) algorithms in CNNs. Following our theoretical results, we propose a SGD algorithm with assurance of almost sure convergence of the methods to a solution at single minimum of classification loss of CNNs. Experimental results show that the proposed method achieves state-of-the-art performance for major image classification benchmarks with CNNs.

\end{abstract}

\section{Introduction}

Over the last decade, convolutional neural networks (CNNs) have been utilized to perform various tasks such as image classification \cite{Alexnet,VGG,image_class} and scene analysis \cite{lecun_scene}. The unprecedented performance of CNNs on these tasks has been attributed, in practice, to design of large-scale datasets, and modeling of more complex architectures with larger number of layers and parameters \cite{Alexnet,go_deeper1}. 


While performing optimization using stochastic gradient descent (SGD) algorithms with backpropagation (BP) in CNNs, we observe that norm of gradients may exponentially increase or decrease \cite{van_grad,exp_grad}. Exploding and vanishing gradients trigger several open problems such as convergence of SGD and its robustness to reparametrization of convolution kernels, and internal covariate shift. In order to cope with these open problems, various normalization methods have been proposed on kernels and/or gradients at initialization, and/or at each update of SGD using different orthogonality constraints \cite{unit_evol,norm_prop,symm_inv,orth1,PRelu,orth2,Lin_2016_CVPR,w_norm,sax,Smith_2016_CVPR}.  However, various kernel normalization methods may affect geometry of search spaces, dissimilarly. More precisely, level sets of classification loss functions, critical points residing in level sets, and convergence properties of SGD may be different for different kernel normalization methods. In order to assure convergence of SGD algorithms to solutions at single minimum, we need to identify search spaces and compute steps according to the geometry of kernel spaces\footnote{In this work, we refer to \textit{convolution kernels} used in CNNs by \textit{kernels}.} in CNNs. In addition, various orthogonality constraints are nonconvex and nonlinear. Therefore, embedding constraints into cost functions may lead to many local minimizers \cite{Wen2013}.

In this work, we address the aforementioned problems by identifying kernel spaces as topological smooth manifolds under a geometric optimization framework for training of CNNs. We pose the kernel estimation problem in CNNs \cite{mnist} as optimization on embedded and/or immersed submanifolds of kernels which are described according to different geometric properties of the kernels, such as orthonormal rectangular or orthogonal square kernels. Thereby, we can define constraints of optimization problems of CNNs in search spaces of SGD algorithms, instead of embedding the constraints into cost functions of the problems \cite{manopt_book,r_precod}. To this end, we first provide theoretical analyses and results to  explore  geometry of kernels in CNNs. Then, we employ our theoretical results for image classification. In our framework, we first construct kernel submanifolds at each layer of a CNN such that a kernel resides as a point on a kernel submanifold. Then, we employ a SGD algorithm for optimization on kernel submanifolds using BP in CNNs. 

\section{Related Work and Summary of Contributions}

Popular kernel normalization methods have been implemented using reparametrizations \cite{data_dep_init,w_norm}, and additional constraints, such as orthogonality \cite{orth1,orth2,path-sgd}, in order to preserve unit norm property of the kernels for forward propagation \cite{norm_prop}, at initialization \cite{init,sax}, or at each epoch of SGD \cite{unit_evol,w_norm}. Unit norm kernels were used for symmetry invariant optimization at the first and second layers of a network in \cite{symm_inv}. One of the challenges of these approaches, besides the lack of theoretical understandings mentioned above, is the employment of the reparametrization and rescaling methods at new layers before/after convolution layers, resulting in an increase of complexity of the network structure by aggregation of the new layers. In addition, statistical properties of data need to be recorded during training, and testing. Therefore, kernel normalization methods may increase computational overhead of CNNs for both training and testing. 

Removal of scale and translation from kernels by normalization can be interpreted as imposition of a geometric structure such that the kernels lie on the sphere  \cite{stoc_geom}. In our approach, embedded kernel submanifolds can be described using the sphere, oblique and/or the Stiefel manifold. Additional constraints can also be imposed using immersed submanifolds such as rotation groups. Thus, our approach can be considered as generalization of the aforementioned approaches such that we can employ our methods to model different submanifolds according to various constraints, such as orthonormal kernels. Thereby, we employ geometry of kernels to identify the constraints on the optimization problem of CNNs. Moreover, gradient descent of natural gradient (NG) methods can be cast as an approximation to SGD for submanifolds which are equipped with Riemannian structure and employed in our framework \cite{sgdman,mirror,smith_2005}. 
In this aspect, our proposed methods can be considered as generalization of NG methods \cite{prong,scaleup_NG}. Our contributions can be summarized as follows:

\noindent \textbf{Analysis of geometry and smooth structures of kernel submanifolds:} One of our main motivations for employment of kernel submanifolds is to assure existence of singular minimum of a loss function $\mathcal{L}$ of CNNs in the search space of a SGD. For this purpose, the loss function is defined as a smooth map $\mathcal{L}: \mathcal{M} \to \mathcal{N}$, where $\mathcal{M}$ is a kernel manifold, and $\mathcal{N}$ is a space of loss values such as a set of classification errors. Thereby, we can formulate the relationship between level sets of $\mathcal{L}$ and submanifolds $\hatM$ of $\mathcal{M}$. We also analyze the conditions under which level sets are submanifolds $\hatM$ that contain critical points  in Section~\ref{sec:geom}. 

\noindent \textbf{A SGD algorithm for optimization on kernel submanifolds in CNNs, and analysis of convergence properties:} By making use of our theoretical results, we propose a SGD algorithm for optimization on kernel submanifolds for training of CNNs in Section~\ref{sec:algorithm}. More precisely, our theoretical results first enable us to employ various smooth manifolds with different metrics to describe submanifolds. For computational efficiency, we then employ Riemannian manifolds to perform steps of SGD methods on submanifolds. We compute steps of SGD according to smooth structures of submanifolds, such as metrics and differential maps, defined on submanifolds, and their topological properties, such as compactness. Moreover, in our proposed SGD algorithm, we can employ momentum and Euclidean gradient decay for optimization on submanifolds extending the methods proposed in \cite{manopt_book,sgdman}. In Section~\ref{sec:conv_results}, we provide two theorems to analyze the convergence of the proposed SGD algorithm. We provide a discussion on computational complexity of the proposed algorithm in Section~\ref{sec:comp}.

To the best of our knowledge, this is the first comprehensive work which employs stochastic optimization methods on embedded and immersed  submanifolds of kernels in CNNs with convergence properties.

The paper is organized as follows. Our proposed mathematical framework is introduced in Section~\ref{sec:geom}. We provide the proposed SGD algorithm and the convergence properties in Section~\ref{sec:algorithm}. In Section~\ref{sec:experiments}, we examine the proposed algorithm, methods and theoretical results for different manifolds using several benchmark datasets in comparison with state-of-the-art methods. Section~\ref{sec:conc} concludes the paper. Proofs of the theorems and implementation details are given in the supplemental material (sup. mat.).

\section{Geometry of Kernel Submanifolds}
\label{sec:geom}

In this work, we contemplate subspaces of convolution kernels endowed with differentiable structures, i.e. kernel submanifolds. Suppose that we are given a set of training samples ${S=\{s_i= (\mathbf{I}_i,y_i) \}_{i=1}^N}$ of a random variable $s$ drawn from a distribution $\mathcal{P}$ on a measurable space $\mathfrak{S}$, where $y_i $ is a class label of the $i^{th}$ image $\mathbf{I}_i$. An $L$-layer CNN consists of a set of tensors $\mathcal{W} = \{\mathcal{W}_l \}_{l=1}^L$, where $\mathcal{W}_l = \{ \mathbf{W}_{d,l} \in \mathbb{R}^{A_l \times B_l \times C_l} \} _{d=1} ^{D_l}$, and ${\mathbf{W}_{d,l} = [W_{c,d,l} \in \mathbb{R}^{A_l \times B_l}]_{c=1}^{C_l}}$  is a tensor\footnote{We use shorthand notation for matrix concatenation such that $[W_{c,d,l}  ]_{c=1}^{C_l} \triangleq [W_{1,d,l}, \cdots,W_{C_l,d,l}]$.} composed of kernels (weight matrices) $W_{d,c,l} $ constructed at each layer $l=1,2,\ldots,L$, for each $c^{th}$ channel $c=1,2,\ldots,C_l$ and each $d^{th}$ kernel $d=1,2,\ldots,D_l$. At each convolution layer, a feature representation $f_l(\mathbf{X}_l;\mathcal{W}_l)$ is computed by compositionally employing non-linear functions, and convolving an image $\mathbf{I}$ with kernels by  
\begin{equation}
f_l(\mathbf{X}_l;\mathcal{W}_{l}) = f_l(\cdot;\mathcal{W}_l) \circ  \cdots \circ f_1(\mathbf{X}_1;\mathcal{W}_{1}),
\label{eq:comp_rep}
\end{equation}
where ${\mathbf{X}_1 := \mathbf{I}}$ is an image for ${l=1}$, and $\mathbf{X}_{l} = [ X_{c,l}]_{c=1}^{C_l}$. The $c^{th}$ channel of the data matrix $X_{c,l}$ is convolved with the kernel ${W}_{c,d,l}$ to obtain the $d^{th}$ feature map ${ X_{c,l+1} : = \hat{X}_{d,l}}$ by $\hat{X}_{d,l} = {W}_{c,d,l} \ast X_{c,l}, \forall c, d, l$ \footnote{We ignore the bias terms in the notation for the sake of simplicity.}. Given a batch of samples $\mathfrak{s} \subseteq S$, we denote a value of a classification loss function for a kernel $\omega \triangleq W_{c,d,l}$ by $\mathcal{L}(\omega,\mathfrak{s})$, and the loss function of kernels $\mathcal{W}$ utilized in the CNN by $\mathcal{L}(\mathcal{W},\mathfrak{s})$. If we assume that  $\mathfrak{s}$ consists of a single sample $s_i$, then, an expected loss or cost function of the CNN  is computed by

\begin{figure}[t!]
	\centering
	\subfloat[Level sets  {\color{Orange} $\hat{\mathcal{M}}$}($\mathfrak{c}$)$\subset \setM$.
	\label{fig:first-case}]{%
		\includegraphics[width=3.4in]{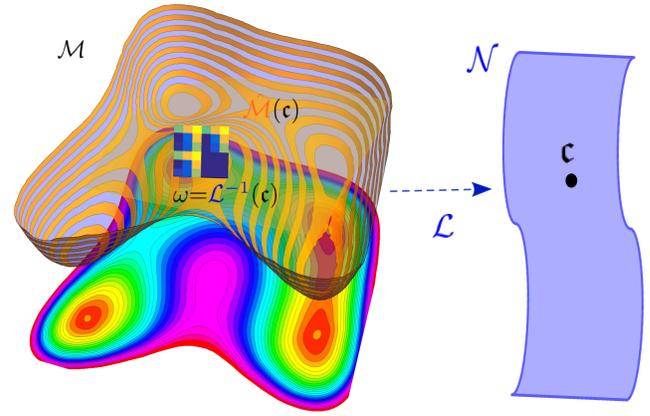}%
	}
	\\
	\subfloat[Zero level sets {\color{Red} $\hat{\mathcal{M}}$}($\mathfrak{0}$).
	\label{fig:second-case}]{%
		\includegraphics[width=3.4in]{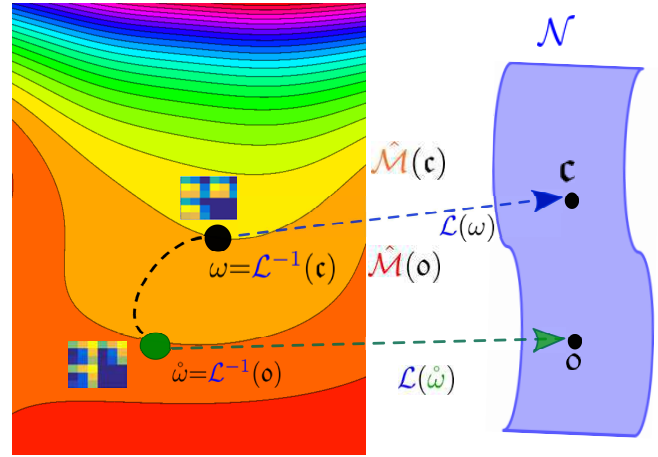}%
	}
	\caption{(a) Level sets  {\color{Orange} $\hat{\mathcal{M}}$}($\mathfrak{c}$)
		residing in kernel spaces $\setM$ induced by a loss function {\color{Blue} ${\mathcal{L}}$} from $\setM$ to a set of error values {\color{Periwinkle} ${\mathcal{N}}$}. (b) Zero level sets {\color{Red} $\hat{\mathcal{M}}$}($\mathfrak{c}=0$) contain critical kernels at which the gradient vanishes (see {Definition~\ref{def:EK}}, and {Theorem~\ref{theorem:rank_dim}}).  }	\label{fig:level_sets}
\end{figure}
\begin{figure}[t]
	
	\centering
	\includegraphics[width=0.475\textwidth]{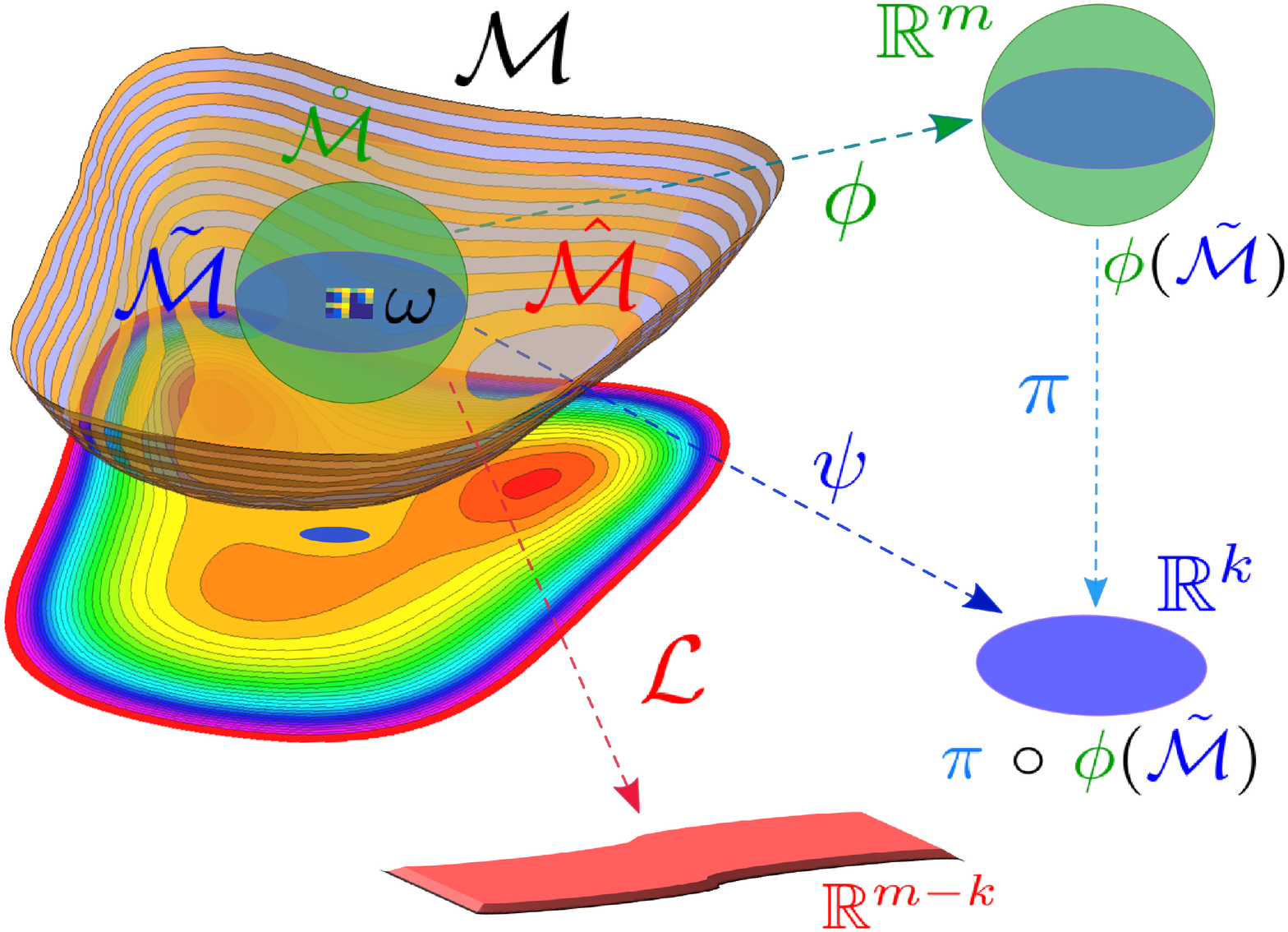}
	\caption{ $\setM$ is a smooth $m$ dim. kernel manifold. {\color{Orange} $\hat{\mathcal{M}}$} is a $k$ dim. submanifold embedded in $\setM$. 
		({$\color{ForestGreen} \mathring{\mathcal{M}}$}
		,{$\color{ForestGreen} \phi$}
		) is a smooth chart of {\color{Orange} $\hat{\mathcal{M}}$} centered at a kernel $\omega \in \setM$.
		${\color{Blue} \tilde{\mathcal{M}}} ={\color{ForestGreen} \mathring{\mathcal{M}}} \cap {\color{Orange} \hat{\mathcal{M}}} $ is a $k$-slice of ${\color{ForestGreen} \mathring{\mathcal{M}}}$ (see {Definition~\ref{def:EK}}). $({\color{Blue} \tilde{\mathcal{M}}, \psi})$ is a chart of $\setM$ centered at $\omega \in \setM$, and {{\color{Cerulean} $\pi$}: {\color{ForestGreen} $\mathbb{R}^m$}{\color{NavyBlue} $\to$}{\color{Blue} $\mathbb{R}^k$}} is the projection onto the first $k$ coordinates. Two charts of $\setM$ and {\color{Orange} $\hat{\mathcal{M}}$} are associated by {\color{Blue} $\psi$}= {\color{NavyBlue} $\pi$} $\circ$ {\color{OliveGreen} $\phi$}({\color{Blue} $\tilde{\mathcal{M}}$}). 
		${\color{Blue} \tilde{\mathcal{M}}} $ is a level set of a loss function ${\color{red} \mathring{\setL}}:  {\color{ForestGreen} \mathring{\mathcal{M}}} \to {\color{red} \mathbb{R}^{m-k}}$ (see {Proposition~\ref{prop:level_sets})}.
	}
	\label{figEK}
\end{figure}

\begin{equation}
\mathcal{L}(\mathcal{W}) \triangleq E_{\mathcal{P}} \{ {\mathcal{L}}(\mathcal{W},s) \} = \int {\mathcal{L}}(\mathcal{W},s) d \mathcal{P}.
\label{eq:expected_cost}
\end{equation}
The expected loss for $\omega$ is denoted by $\mathcal{L}(\omega)$. For a finite set of samples $S$,  $\mathcal{L}(\mathcal{W})$ is approximated by an empirical loss $\frac{1}{|S|} \sum_{i=1}^{|S|} \mathcal{L}(\mathcal{W},s_i)$, where $|S|$ is the size of $S$ (similarly, $\mathcal{L}(\omega)$ denotes the empirical loss for $\omega$). Then, feature representations are learned using SGD by solving
\begin{equation}
\min_{\mathcal{W}} \mathcal{L}(\mathcal{W}).
\label{eq:opt1}
\end{equation}
Restriction on the search space can be imposed into \eqref{eq:opt1} by defining constraints expressed as a function of the variables $\mathcal{W}$, such as normalization \cite{path-sgd}. If the search space is equipped with a manifold structure, then  constrained optimization problem  of CNNs can be converted into that of unconstrained optimization. We explore the geometric relationship between the loss \eqref{eq:expected_cost} and kernels, by first defining 
\begin{equation}
\hat{\mathcal{M}} (\mathfrak{c} ) \triangleq \{ \omega \in \mathcal{M}:   {\mathcal{L}}(\omega) = \mathfrak{c}, \mathfrak{c} \in \mathcal{N} \}   ,
\label{eq:variety}
\end{equation}
where the loss function $\mathcal{L}$ is a map from a kernel manifold $\mathcal{M}$ to a set of classification errors $\mathcal{N}$ (see Fig.~\ref{fig:level_sets}). For example, $\mathcal{L}$ can be considered as a classification loss that maps kernels residing in a subspace of $A \times B$ matrices $\mathbb{R}^{A \times B}$ to a subspace of $\mathbb{R}$. Thus, $\mathcal{M}$ is partitioned into level sets $\hat{\mathcal{M}} (\mathfrak{c} ) $, $ \forall \mathfrak{c}$. A gradient computed at a kernel $\omega \in \mathcal{M}$, denoted by  $\gr \mathcal{L}(\omega)$ is orthogonal to the level set at $\omega$. The gradient  vanishes for a critical value $\mathfrak{c} = \mathbf{0}$ at $\hat{\omega} \in \mathcal{M}$, called a \textit{critical kernel}.  By the Weierstrass' theorem, if $\mathcal{L}$  is a continuous function, and $\setM$ is a closed and bounded (compact)
manifold, then $\setM$ has a minimum  \cite{Luenb}. Then, there exists a closed subset ${\bar{\mathcal{M}} \subset \mathcal{M}}$  such that ${\mathcal{L}}^{-1} (\mathbf{0}) =  \bar{\mathcal{M}}$, if $\mathcal{L}$ is a loss function of class $C^{\infty}$.	Fortunately, several popular loss functions, such as exponential and sigmoid loss, are $C^{\infty}$. We use this property by assuming that $ \omega \triangleq {W_{c,d,l} \in \mathbb{R}^{A_l \times B_l}}, \forall c,d$, computed at the $l^{th}$ layer reside in submanifolds of smooth topological manifolds \cite{lee_smooth}. %
If $\bar{\mathcal{M}}$ is a submanifold, then the notion of critical kernel is equivalent to that of extrema. This result motivates us to analyze and employ the geometry of kernel submanifolds while solving \eqref{eq:opt1} using SGD. Following this motivation, we introduce a procedure to describe and construct embedded submanifolds of $\mathcal{M}$. If $\openM$ is an open subset of $\mathbb{R}^m$, then a $k$-slice of $\openM$ is any subset 
\begin{equation}
\mathfrak{C}=  \{ (\nu_1, \ldots, \nu_m) : \nu_i = \mathfrak{c}_i,  i =k+1,\ldots,m, k\geq 1   \},
\label{eq:slice}
\end{equation}
where $(\nu_1, \ldots, \nu_m )$ are called slice coordinates of $\mathfrak{C}$ for constants $\mathfrak{c}_i$ \cite{lee_smooth}. Next, we define embedded submanifolds.
\begin{definition}[Embedded kernel submanifolds]
	\label{def:EK}
	
	Suppose that $\setM$ is a smooth $m$ dimensional (dim.) kernel manifold, and ${\color{Orange} \hat{\mathcal{M}}}
	\subset \setM$. Let ${\color{OliveGreen} \phi}: {\color{ForestGreen} \mathring{\mathcal{M}}} \to \openM'$ be a one-to-one and onto function with a continuous inverse which maps ${\color{ForestGreen} \mathring{\mathcal{M}}} \subset \setM$ containing a kernel $\omega$ to an open set $\openM' \subseteq \mathbb{R}^m$. Suppose that, for each  $\omega \in {\color{Orange} \hat{\mathcal{M}}} $, there exists a tuple $({\color{ForestGreen} \mathring{\mathcal{M}}}, {\color{OliveGreen} \phi})$ for $\setM$ such that ${\color{Blue} \tilde{\mathcal{M}}} ={\color{ForestGreen} \mathring{\mathcal{M}}} \cap {\color{Orange} \hat{\mathcal{M}}} $ is a $k$-slice of ${\color{ForestGreen} \mathring{\mathcal{M}}}$. Then, ${\color{Orange} \hat{\mathcal{M}}}$ is called a $k$ dimensional \textit{embedded kernel submanifold} of $\setM$  (see Fig.~\ref{figEK}). \QEDbs	
\end{definition}

{Definition~\ref{def:EK}} describes a relationship between partitioning of $\setM$ into level sets that contain critical kernels using \eqref{eq:variety} (see Fig.~\ref{fig:level_sets}), and into embedded kernel submanifolds that satisfy \eqref{eq:slice} (see Fig.~\ref{figEK}). In other words, we consider an approach to construct embedded kernel submanifolds which correspond to level sets of loss functions on manifolds that contain critical kernels. However, not every level set corresponds to an embedded kernel submanifold. In the next theorem, we introduce a condition to associate a level set to an embedded kernel submanifold.
\begin{theorem}[Conditions required to identify level sets by embedded kernel submanifolds]
	If a loss ${\color{blue} \setL}: \mathcal{M} \to {\color{blue}\mathcal{N}}$ is a smooth map with constant rank $r$ (i.e. the rank of the Jacobian matrix of ${\color{blue} \setL}$), then each level set of ${\color{blue} \setL}$ is an embedded kernel submanifold ${\color{red}\hat{\mathcal{M}}} \subseteq \mathcal{M}$ (see Fig.~\ref{fig:level_sets}). 
	\label{theorem:rank_dim}
\end{theorem}

In addition, not all embedded kernel submanifolds can be expressed as level sets of a loss $\mathcal{L}$ that contain critical kernels even if the loss is a smooth submersion.
However, the next proposition shows that every embedded kernel submanifold can be
locally expressed as a level set. 

\begin{prop}[Conditions required to assure that embedded kernel submanifolds contain zero sets]
	Let ${\color{Orange} \hatM}$ be a subset of an $m$ dim. smooth kernel manifold $\setM$. Then, ${\color{Orange} \hatM}$ is an embedded kernel submanifold of $\setM$ and its dimension is $k$ if and only if every kernel $\omega \in {\color{Orange} \hatM}$ has a neighborhood
	${\color{ForestGreen} \mathring{\mathcal{M}}}$ in $\setM$ such that ${\color{Blue} \tilde{\mathcal{M}}} ={\color{ForestGreen} \mathring{\mathcal{M}}} \cap {\color{Orange} \hat{\mathcal{M}}} $ is a level set of a loss ${\color{red} \mathring{\setL}}:  {\color{ForestGreen} \mathring{\mathcal{M}}} \to {\color{red} \mathbb{R}^{m-k}}$ which is a smooth submersion.	The level set contains a zero set $\hat{\mathcal{M}} (0 )$ (see Fig.~\ref{fig:level_sets} and  Fig.~\ref{figEK}).  
	\label{prop:level_sets}
\end{prop}

In order to perform optimization on kernel submanifolds whose inclusion maps are injective submersions, such as rotation groups, we need to consider a more general notion of submanifolds characterized by immersed kernel submanifolds (see the sup. mat. for details). A Lie subgroup of a Lie group $\mathcal{G}$, e.g. the rotation group, is endowed
with a topology and smooth structure making it into a Lie group and an immersed kernel submanifold of $\mathcal{G}$. Consequently, embedded kernel submanifolds, which are also subgroups of $\mathcal{G}$, are automatically Lie subgroups. Therefore, our framework enables us to employ both embedded and immersed kernel submanifolds in SGD. In practice, this property is required by the SGD algorithms in order to train CNNs using various normalized kernels including orthogonal square kernels with determinant 1 (i.e. members of the rotation group) with assurance of convergence. Next, we use our framework to explore geometry of space of normalized kernels.

\subsection{Geometry of Space of Normalized Kernels}
\label{sec:theory_analysis}
Analysis of the geometry of submanifolds of normalized kernels is an open and understudied problem. This is a crucial problem since gradient steps should be identified by the geometry of submanifolds as explained in the previous sections. In the next theorem, we explore this conjecture using concrete examples for different normalization methods. 

\begin{prop}[Geometry of submanifolds of normalized kernels]
	Suppose that we are given a set of kernels 	$ { \omega \triangleq {W_{d,c,l} \in \mathbb{R}^{A_l \times B_l}} }, \forall c,d$ computed at the $l^{th}$ layer of a CNN. Moreover, suppose that an ambient manifold $\setM$ is identified by a Euclidean space of $A \times B$ matrices $\mathbb{R}^{A \times B}$.
	
\noindent	\textbf{(i)} If the kernels $\omega$ are normalized to the unit Frobenius norm, then each kernel resides on the $A_lB_l-1$ dimensional sphere ${\mathcal{S}(A_l,B_l) =  \{\omega \in \mathbb{R}^{A_l \times B_l} : \| \omega \|^2_{F} = 1\}}$, where $\| \cdot \|_F$ is the squared Frobenius norm.
	
\noindent	\textbf{(ii)} If the columns $\omega_b, \forall b=1,2,\ldots,B_l$ of $\omega$ are normalized with the unit norm, then each $\omega_b$ resides on the sphere $ \mathcal{S}(A_l) =  \{\omega_b \in \mathbb{R}^{A_l} :  \| \omega_b \|^2_{F} = 1\}$. Then, the space of  $\omega$ is isometric to the oblique manifold  ${ \mathcal{OB}(A_l, B_l)= \{\omega \in \mathbb{R}^{A_l \times B_l} : {\rm ddiag} (\omega ^{\rm T} \omega) = I_{B_l}\}}$, and ${\rm ddiag}(\omega)$ denotes the diagonal matrix whose diagonal elements are those of $\omega$, and $I_{B_l}$ is a $B_l \times B_l$ identity matrix.
	
\noindent	\textbf{(iii)} If the kernels are orthonormal, then they reside on the compact Stiefel manifold $ {St(A_l, B_l) = \{\omega \in \mathbb{R}^{A_l \times B_l} : \omega^T \omega = I_{B_l}\} }$. If the shapes of orthonormal kernels are square  such that ${A_l= B_l=n}$, then they reside on the orthogonal group ${\mathcal{O}(n) = \{\omega \in \mathbb{R}^{n \times n} : \omega^T\omega = I_n \}}$. Moreover, if $\det(\omega) =+1$, $\forall \omega$, then the kernels reside in the rotation group.
	
	\label{corr:corr_norm1}
\end{prop}

This theorem shows that different kernel normalization methods, even if they are implemented in an intuitively similar manner, imply different geometric properties. For instance, if the kernels are normalized to the unit Frobenius norm, then they reside on the $A_lB_l-1$ dimensional sphere. Besides, if the kernels are first orthonormalized as suggested in (iii), then each column of the kernel also resides on the sphere $\mathcal{S}(A_l)$. However, the kernel $W_{d,c,l}$ resides on the Stiefel manifold. In addition, if the shape of  $W_{d,c,l}$ is a square, then kernels reside on $\mathcal{O}(n)$. If each column is first normalized with unit norm, then $W_{d,c,l}$ resides on a kernel submanifold isometric to $\mathcal{OB}(A_l, B_l)$. 

In SGD, we need to pay attention to restriction of gradients and steps employed on kernel submanifolds to assure convergence to a solution. The first reason is that kernels should reside in \textit{locally} compact sets to assure existence of critical kernels (see {Theorem~\ref{theorem:rank_dim}} and {Proposition~\ref{prop:level_sets}}). Second, while performing SGD steps, gradients of kernels should be bounded in the compact sets, which can be achived by employing mappings from Euclidean gradients obtained using BP to submanifold gradients residing on tangent spaces of kernel submanifolds. However, these two requirements are not considered in the state-of-the-art normalization methods, resulting in exploding and vanishing gradients. Note that compact sets and mappings of gradients are computed according to manifold and smooth structures of submanifolds. Therefore, we need to employ appropriate mappings of kernels and gradients in SGD while performing steps. In order to perform SGD for different kernel submanifolds assuring almost sure convergence to a solution, we suggest a SGD algorithm considering an optimization approach on Riemannian manifolds in the next section. 

\section{A SGD Algorithm for Optimization on Kernel Submanifolds in CNNs}
\label{sec:algorithm}

Various optimization algorithms have been developed to solve optimization problems on matrix manifolds \cite{manopt_book,manopt}. However, development of SGD algorithms on kernel submanifolds, and analysis of their convergence properties in CNNs have not been addressed yet. In our framework, we perform optimization on kernel submanifolds at each convolution layer of an $L$-layer CNN. An algorithmic description of our proposed methods is given in {Algorithm~\ref{alg1}}:


\noindent $\bullet$ \textbf{Initialization:} We first define a KS  $\hatM_l$, for each convolution layer $l=1,2,\ldots,\mathcal{L}$ whose members are $ {\omega_l^t \in \hatM_l}$, where $ \omega_l^t \triangleq {W}_{d,c,l}^t$, $\forall c=1,2,\ldots,C_l$, $\forall {d=1,2,\ldots,D_l}$, $\forall l$.

\begin{figure}[t]
	
	\centering
	\includegraphics[width=0.475\textwidth]{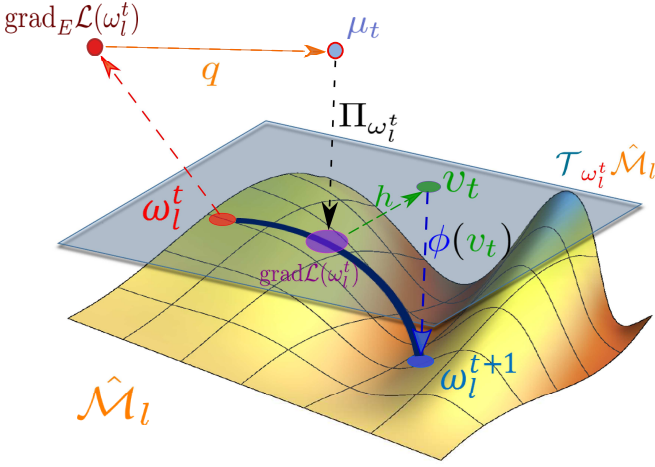}%
	\caption{Updating kernels on a kernel submanifold ${\color{BurntOrange} \hat{\mathcal{M}}_l}$, $\forall l$ (Lines 4-9 in the {Algorithm~\ref{alg1}}) at the $t^{th}$ epoch of a SGD. }
	\label{fig_updates}
\end{figure}

\noindent $\bullet$ For each epoch $t =1,2,\ldots,T$, and for each $l=1,2,\ldots,{L}$, following steps are performed (see Fig.~\ref{fig_updates});

\textbf{- Line 4:} The Euclidean gradient $\gr_E \mathcal{L}(\omega_l^{t})$ is computed and obtained using backpropagation  (see Fig.~\ref{fig_updates}).  

\textbf{- Line 5:} Momentum and Euclidean gradient decay methods are employed on the Euclidean gradient $\gr_E \mathcal{L}(\omega_l^{t})$ using $\mu_t :=  q \Big ( \gr_E \; \mathcal{L}(\omega_l^{t}),\mu_t,\Theta \Big)$ (see Fig.~\ref{fig_updates}). We can employ state-of-the-art acceleration methods \cite{on_mom} modularly in this step. For example, momentum can be employed with the Euclidean gradient decay using
\begin{equation}
	q \Big ( \gr_E \; \mathcal{L}(\omega_l^{t}),\mu_t,\Theta \Big) = \theta_{\mu} \mu_t - \theta_E \gr_E \; \mathcal{L}(\omega_l^{t}),
	\label{mom_decay}	
\end{equation} 
where $\theta_{\mu} \in \Theta$ is the parameter employed on the momentum variable $\mu_t$. We consider $\theta_E \in \Theta$ as the decay parameter for the Euclidean gradient. The reason is that $\theta_{\mu}$ and $\theta_E$ affect the step performed in the ambient Euclidean space while the learning rate (LR) is employed on the submanifold gradient. A detailed discussion of methods that are used to compute $q(\cdot)$ is given in  the sup. mat.     

\begin{algorithm}[t]
	\SetAlgoLined

	\KwInput{$T$ (number of iterations), $S$ (training set), \\ $\Theta$ (set of hyperparameters) and $\mathcal{L}$ (a loss function). }
	\textbf{Initialization:} Construct kernel submanifolds~$\{ \hatM_l \}_{l=1}^{ L}$, and initialize
	$ \omega_l^t \in \hatM_l$, where $ \omega_l^t \triangleq {W}_{d,c,l}^t$, $\forall c, d,l$. 
	
	\For{each iteration $t=1,2,\ldots,T$}{
		
		\For{each layer $l=1,2,\ldots,{L}$}{
			
			Compute the Euclidean gradient $\gr_E \mathcal{L}(\omega_l^{t})$.
			
			$
			\mu_t :=  q \Big ( \gr_E \; \mathcal{L}(\omega_l^{t}),\mu_t,\Theta \Big)$.
			
			$
			\gr \mathcal{L}(\omega_l^{t}) := {\rm \Pi}_{\omega_l^t}  \mu_t$.

			$\alpha_t :=g(t,\Theta) $.
			
			$ v_t := h(\gr \mathcal{L}(\omega_l^{t}), \alpha_t)$.
			
			$
			\omega_l^{t+1} := \phi_{\omega_l^t}(  v_t), \forall \omega_l^t \in \hatM_l.
			$
			
			$ t := t+1$.
			
		}
	}
	\KwOutput{Set of estimated kernels $\{\omega_l^T \}_{l=1}^{{L}}$. }
	\caption{SGD on kernel submanifolds.}
	\label{alg1}
	
\end{algorithm}
   
\textbf{- Line 6:} The moved vector $\mu_t$ is projected to the tangent space $\mathcal{T}_{\omega_l^t}\hatM_l$ at $\omega_l^t$,  to compute the submanifold gradient $
\gr \mathcal{L}(\omega_l^{t}) := {\rm \Pi}_{\omega_l^t} \mu_t$,
where ${\rm \Pi}_{\omega_l^t}$ is a projection operator defined according to the geometry of $\hatM_l$  (see Fig.~\ref{fig_updates}), and is used to bound the norm of the gradient.

\newpage
\textbf{- Line 7:} The learning rate $\alpha_t$ is updated by ${\alpha_t :=g(t,\Theta)}$, where  $g(t,\Theta)$ is a function that controls the convergence rate \cite{sgdman}. We choose $g(t,\Theta)$ which satisfies the following as suggested in \cite{sgdman,kiefer1952,lecun-efficient-backprop-1998};
\begin{equation}
\sum_{t=0} ^{\infty} g(t,\Theta) = +\infty \; {\rm and} \; \sum_{t=0} ^{\infty} g(t,\Theta)^2 < \infty.
\label{eq:rate}
\end{equation} 

\textbf{- Line 8:} A vector $v_t \in \mathcal{T}_{\omega_l^t}\hatM_l$ is computed using $ {v_t := h(\gr \mathcal{L}(\omega_l^{t}), \alpha_t)}$, where $h(\cdot)$ is a function that defines the next step on the tangent space $\mathcal{T}_{\omega_l^t}\hatM_l$ at $\omega_l^t$ (see Fig.~\ref{fig_updates}). In this work, we employed $	{h(\gr \mathcal{L}(\omega_l^{t}), \alpha_t) := -  \alpha_t \gr \mathcal{L}(\omega_l^{t})}$ to move the solution in a gradient descent direction with step size $\alpha_t$.

\textbf{- Line 9:} Compute the next iterate using ${\omega_l^{t+1} := \phi_{\omega_l^t}(  v_t)}, \forall \omega_l^t \in \hatM_l$, where $\phi$ is a mapping from $\mathcal{T}_{\omega_l^t}\hatM_l$ onto $\hatM_l$ (see Fig.~\ref{fig_updates}). We employ exponential maps and/or retractions for implementation of $\phi$ (see the supp. mat. for details). Therefore, this step enables us also to keep $\omega_l^{t+1}$ on a \textit{locally compact} subset of the kernel submanifold $\hatM_l$, $\forall l$.

\subsection{Convergence properties of {Algorithm~\ref{alg1}}}
\label{sec:conv_results}

In \cite{sgdman}, a procedure was developed to analyze convergence properties of SGD methods for a particular class of manifolds following the proof methods suggested in \cite{saadonline}. In this work, we extend and employ their results to train CNNs using  different kernel submanifolds. We first consider a collection of kernels $\{ \omega_l^t \}_{t\geq1} $ computed at the $l^{th}$ layer of a CNN as a stochastic process. Then, the expected value of a submanifold gradient of a classification loss can be computed by $
\nabla \mathcal{L}(\omega_l^t) =  E_{\mathcal{P}} \{ \gr {\mathcal{L}}(\omega_l^t,s)  \}$\footnote{In practice, we receive a batch of samples $\mathfrak{s}^t \subseteq S$ at each $t^{th}$ epoch. Assuming that each batch contains a single sample, $\nabla \mathcal{L}(\omega_l^t)$ denotes an average gradient computed by $\frac{1}{|S|} \sum_{i=1}^{|S|} \gr {\mathcal{L}}(\omega_l^t,s_i)$.}. In the following theorems, we provide  convergence properties of {Algorithm~\ref{alg1}} for two cases where we use i)  exponential maps, and ii) retractions for $\phi$  at the $9^{th}$ step of the algorithm.

\textbf{i) Exponential maps for $\phi$ :} An exponential map is used to map a vector $v_t \in \mathcal{T}_{\omega_l^t}\hatM_l$ to a kernel along a geodesic curve on $\mathcal{M}_l$ which goes through $\omega_l^t$ in the direction of $v_t$.
\begin{theorem}[Convergence of {Algorithm~\ref{alg1}} using exponential maps for $\phi$]
	Suppose that the following conditions are satisfied;
	
\noindent \textbf{(1) \label{cond1} Condition for maps onto kernel submanifolds:} $\hatM_l$ is a connected compact Riemannian kernel submanifold.

\noindent \textbf{(2) \label{cond2} Condition for kernels:} There exists a compact set $\mathcal{K}$ such that $ \omega_l^t \in \mathcal{K}$, $\forall t \geq 1$. The minimal distance  between conjugate kernels $\omega_l^t$ and $\omega_l^{t+1}$ denoted by $\rho(\omega_l^t,\omega_l^{t+1})$ is a geodesic that satisfies $\rho(\omega_l^t,\omega_l^{t+1})>0,\forall t,l$.
		
\noindent \textbf{(3) \label{cond3} Condition for gradients:} The gradient is bounded on $\mathcal{K}$, such that $\exists \mathfrak{K}>0$, $\| \gr \mathcal{L}(\omega_l^{t},s) \| \leq \mathfrak{K}$, $\forall s$ and $\forall \omega_l^t \in \mathcal{K}$.
		
\noindent \textbf{(4) \label{cond4} Condition for the classification loss function}:  We use a three times continuously differentiable function $\mathcal{L}(\omega_l) \geq 0$.
		
	 Then, the loss function and the gradient converges almost surely (a.s.) by $\mathcal{L}(\omega^t_l) \xrightarrow[t \to \infty]{\rm a.s.} \mathcal{L}(\hat{\omega}_l)$,
	where $\hat{\omega}_l$ is a minimum, and $\nabla \mathcal{L}(\omega^t_l) \xrightarrow[t \to \infty]{\rm a.s.} 0$.
	\label{thm:exp_conv}	
\end{theorem}

\textbf{ii) Retractions for $\phi$ :} In the experimental analysis, we used retractions to compute numerical approximations to exponential maps onto kernel submanifolds \cite{manopt_book}. Thus, we next provide the convergence properties for the case where $\phi$ is implemented using retractions.
\begin{theorem}[Convergence of {Algorithm~\ref{alg1}} using retraction for $\phi$]
	
	Suppose that the conditions \textbf{(1)-(4)} of {Theorem~\ref{thm:exp_conv}} are satisfied, and that $\phi$ is a twice continuously differentiable retraction. Then, we have $
	\mathcal{L}(\omega^t_l) \xrightarrow[t \to \infty]{\rm a.s.} \mathcal{L}(\hat{\omega}_l)$ and $\nabla \mathcal{L}(\omega^t_l) \xrightarrow[t \to \infty]{\rm a.s.} 0$.
	\label{thm:retr_conv}
\end{theorem}

\subsection{Computational complexity of {Algorithm}~\ref{alg1}} 
\label{sec:comp}

Compared to traditional SGD algorithms \cite{res_net,nature_deep}, the computational complexity of {Algorithm}~\ref{alg1} is dominated by computation of the maps $\Pi$ and $\phi$ at line 6 and 9, depending on the structure of the kernel submanifold used in the algorithm at the $l^{th}$ layer. Concisely, the computational complexity of $\Pi$ is determined by computation of different norms that identify the submanifolds. For instance, for the sphere, we use $\Pi_{\omega_l^t} \mu_t \triangleq (1- \| \omega_l^t \|_F^2) \mu_t$. Thereby, for an $A \times A$ kernel, the complexity is bounded by $\mathcal{O}(A^3)$. Similarly, the computational complexity of $\phi$ depends on the submanifold structure. For example, the exponential maps on the sphere and oblique manifold can be computed using functions of $\sin$ and $\cos$ functions, while that on the Stiefel manifold is a function of matrix exponential. For computation of matrix exponential, various numerical approximations with $\mathcal{O}(\epsilon A^3)$ complexity were proposed for different approximation order $\epsilon$ \cite{fisk,Higham,kenney,nineteen}. However, unit norm matrix normalization is used for computation of retractions on the sphere and the oblique manifold. Moreover, QR decomposition of matrices is computed with $\mathcal{O}(A^3)$ \cite{golub} for retractions on the Stiefel manifold. In addition, the computation time of maps can be reduced using parallel computation methods. For instance, a rotation method was suggested to compute QR using $\mathcal{O}(A^2)$ processors in $\mathcal{O}(A)$ unit time in \cite{fast_qr}. Therefore, computation of retractions is computationally less complex compared to the exponential maps. Since the complexity analysis of these maps is beyond the scope of this work, and they provide the same convergence properties for our proposed algorithm, we used the retractions in the experiments. Detailed formulations of maps, and implementation details are given in the sup. mat.

\newpage
\section{Experimental Analyses and Results}
\label{sec:experiments}


The proposed framework and the algorithm can be employed to train CNNs using different manifolds. We train state-of-the-art CNNs using our algorithm on benchmark datasets by optimization on three kernel submanifolds, namely the sphere, the oblique and the Stiefel manifold. In order to perform a fair performance comparison, we used the same code and hyperparameters provided by the authors. 
Implementation details are given in the sup. mat. 


 We can explore the relationship between features learned using different manifolds by expounding geometry of manifolds of kernels $ \{ \mathbf{W}_{d,l} \in \mathbb{R}^{A_l \times B_l \times C_l} \} _{d=1} ^{D_l}, \forall l=1,2,\dots,L,$ computed at convolution layers and fully connected (FC) layers of a CNN consisting of $L$ layers. We first note that, in our framework, we identify KMs by kernels $ \mathbf{W}^{fc}_{l} \in \mathbb{R}^{C_l \times D_l } $  at FC layers, since $A_l=B_l=1$ for FC layers. Then, we delineate the structure of patterns learned at classification layer and the lower layers according to the constraints imposed on kernels by the manifolds (see {Proposition~\ref{sec:theory_analysis}}):

\noindent	$\bullet$ \textbf{At the classification layers} ($l=L$), constraints imposed on FC kernels by manifold structures enable us to perform regularization \cite{orth}. Thereby, our proposed framework and methods enable us to explore and utilize the relationship between two properties of regularization methods, namely i) regularization of models using data augmentation \cite{dataAugReg}, and ii) learning of models endowed with geometric invariants \cite{RegInv}. For instance, the Stiefel manifold implies an $\ell_2$ norm regularization for classification \cite{Bak} as utilized in ridge regression \cite{lasso}. Since $\ell_2$ regularized logistic regression is rotationally invariant  \cite{RegInv}, we perform  regularization using the Stiefel equivalent to the regularization obtained by generating rotated samples in data augmentation \cite{dataAugReg}. Moreover, compact class conditional probability density functions can be learned over the Stiefel manifold \cite{Turaga1,Turaga2}. If the sphere is used, then we perform trace norm regularization \cite{Trnorm2,Trnorm} since kernels residing on the sphere are normalized using the Frobenius norm \cite{golub}. Since generalized trace norms can be considered as the analog for matrices of what the weighted $\ell_1$ norm is for vectors \cite{Trnorm3}, we perform regularization using kernels of the sphere as performed by Lasso type algorithms \cite{lasso}. Moreover, we perform regularization on off-diagonal elements of kernels of the oblique manifold by assuming independence between covariates \cite{offdiag2,offdiag1,offDiag,Pourahmadi}. 

\noindent	$\bullet$ \textbf{At the lower layers} ($l <L$), if we use kernels with $A_l >1$ and $B_l >1$, then we perform additional regularization on spatially distributed patterns \cite{disc_shape,spt_reg} within a neighborhood determined by $A_l$ and $B_l$. For instance, square shape kernels of the Stiefel manifold construct the orthogonal group by {Proposition~\ref{sec:theory_analysis}}. Then, the orthogonality constraints imposed by these kernels enable us to learn representations of shape variation caused by both shape deformation and viewpoint changes \cite{StiefelShape2,StiefelShape1}. Moreover, translation and scaling variability is removed from kernels of the sphere. This property has been used for statistical shape analysis to learn representations of unit length curves, shape primitives \cite{mozay_iccv15,SpShape}, and deformable shapes \cite{Grenander_pt}. Therefore, these shape representations can be learned by training CNNs using kernels on the sphere. Moreover, the constraints determined by the oblique manifold induce oblique rotation. Thereby,  features which are mutually independent, and the orthogonal transformations that minimize the dependence between features, can be learned using the kernels on the oblique manifold\cite{oblq}.

Since a detailed analysis of each of these properties is beyond the scope of this work, we explore them experimentally by training different CNNs with the aforementioned manifolds, and analyzing their performance.

\subsection{Comparison with Normalization Methods}

In a recent work \cite{w_norm}, kernels are reparameterized by a fixed norm $r$ that is initialized by the inverse of standard deviation of pre-activations. Thereby, their proposed method constructs a space of kernels identified by the sphere with radius $r$. In this aspect, their proposed method can be considered as a realization of our proposed algorithm for the sphere. In other words, we can perform optimization on other kernel manifolds such as the oblique and the Stiefel manifold in addition to the sphere. Moreover, each kernel can reside in a different manifold endowed with a different geometry. For instance, we can perform optimization on different kernels that reside on the spheres with different radii. Therefore, our proposed methods enable us to have a better control on the geometry of kernel spaces for training of CNNs compared to their method \cite{w_norm}. 

We examine this property by training their proposed CNN architecture \cite{w_norm}, which is a variation of All-CNN architecture and denoted by SK, using our proposed methods for the sphere (Sp), the oblique manifold (Ob) and the Stiefel manifold (St). The results given in Table~\ref{tab:cifar10woDA} are obtained by training CNNs on the Cifar-10 dataset without using data augmentation (DA). In Table~\ref{tab:cifar10woDA}, we observe that our methods that use the sphere (SK $\dagger$ (Sphere)) outperform the methods proposed in  \cite{w_norm} (SK $\dagger$ (WN)). This observation supports our claim for the benefit of employment of kernels belonging to spaces with various manifold structures, e.g. the sphere with varying radii (which are determined by statistical properties of the data and the gradients of the loss). We also observe that we further boost the performance using the oblique and the Stiefel manifolds. This result also propounds conjectures and results provided in the previous works \cite{Lui12,minh2016algorithmic,turaga2015riemannian} regarding regularization and invariance properties of models learned using different manifolds.

\begin{table}[t]
	\centering
	\caption{Results for Cifar-10 without DA. The results marked by $\dagger$ indicate the results reproduced by our implementation of  the associated algorithm using the code provided by the authors of the related work. Classification error obtained using the baseline CNN is marked by {\color{red} red}, and our best error is marked by {\color{blue} blue}. }
	\begin{tabular}{C{4.8cm} C{2.65cm}}
		\toprule
		\toprule
		\textbf{Model} & \textbf{Class. Error (\%)} \\
		\midrule
		\midrule
		NiN \cite{nin} / NiN $\dagger$ & 10.41 / {\color{red} 10.68}\\
		NiN + MOBN (Sp / Ob / St) $\dagger$ & 9.03 / 8.95 / {\color{blue} \textbf{8.57}}\\
		NormProp \cite{norm_prop} / All-CNN-C \cite{Allconv} & 9.11 / 9.08\\
		SK \cite{w_norm}/ SK $\dagger$ / SK + MOBN \cite{w_norm}& 8.43 / {\color{red}8.45} /8.52\\
		SK (WN) \cite{w_norm} / SK (WN) $\dagger$ & 8.46 /  8.51\\
		SK (Sp / Ob / St) $\dagger$  & 8.24 / 8.11 / {\color{blue} \textbf{7.94} }\\
		SK (BN) \cite{w_norm} & 8.05\\
		SK + MOBN (WN) \cite{w_norm} /$\dagger$ & 7.31 / {\color{red} 7.33}\\
		SK + MOBN (Sp / Ob / St) $\dagger$ & 6.88 / 6.75 / {\color{blue} \textbf{6.02} }\\
		\bottomrule
		\bottomrule
	\end{tabular}%
	\label{tab:cifar10woDA}%
\end{table}%

\begin{table*}[ht]
	\centering
	\caption{Classification error (\%) for larger networks trained on Cifar-10 and Cifar-100 datasets with and without using DA.}
	\begin{tabular}{C{4.199cm}|C{2.37cm}|C{2.890cm}|C{2.89cm}|C{2.89cm}|}
		\toprule
		\toprule
		\multicolumn{1}{c}{\textbf{Model}} & \multicolumn{1}{c}{\textbf{Cifar-10 w. DA}} & \multicolumn{1}{c}{\textbf{Cifar-100 w. DA}}  & \multicolumn{1}{c}{\textbf{Cifar-10 w/o DA}} & \multicolumn{1}{c}{\textbf{Cifar-100 w/o DA}} \\
		NormProp \cite{norm_prop} & 7.47 & 29.24  & 9.11 & 32.19 \\
		PRONG (8 conv. layers) \cite{prong} & 7.32 & -  & - & - \\
		RCD \cite{DCCN} / RCD \cite{SN} / RCD$\dagger$ & 6.41 / $\sim$ / {\color{red} 6.58} & 27.22 / 27.76 / {\color{red} 27.52}        & 13.63 / $\sim$ / {\color{red} 13.60} & 44.74 / $\sim$ / {\color{red} 45.09} \\
		RCD + MOBN (Sp/Ob/St) $\dagger$ & 6.22 / 6.07 / {\color{blue} \textbf{5.93}} & 26.44 / 25.99 / {\color{blue} \textbf{25.41}} & 13.11 / 12.94 / {\color{blue} \textbf{12.88}} & 42.51 / 42.30 / {\color{blue} \textbf{40.11}} \\
		RSD \cite{DCCN} / RSD \cite{SN} / RSD  $\dagger$  & 5.23 / 5.25 / {\color{red} 5.63} & 24.58 / 24.98 / {\color{red} 25.03} & 11.66 / $\sim$ / {\color{red} 11.68} & 37.80 / $\sim$ / {\color{red} 38.15} \\
		RSD + MOBN (Sp / Ob / St)$\dagger$ & 5.20 / 5.14 / {\color{blue} \textbf{4.79}}  & 23.77 / 23.81 / {\color{blue} \textbf{23.16}} & 10.91 / 10.93 / {\color{blue} \textbf{10.46}} & 36.90 / 36.47 / {\color{blue} \textbf{35.92}} \\
		\bottomrule
		\bottomrule		
	\end{tabular}%
	\label{tab:rcd}%
\end{table*}%

We aim to learn representations robust to statistical variance and mean of pre-activations by normalizing them using batch normalization (BN). If features input to a layer are i.i.d. with zero mean and unit variance, then we can equivalently obtain these robust representations using  normalized kernels at that layer \cite{w_norm}. This property is explored in \cite{NormProp} by proving that approximation error to covariance of pre-activations is upper bounded by a function of kernel norms. Therefore, normalized kernels that reside in the sphere are used to train CNNs in \cite{NormProp}. Following this property, BN is employed by removing just mean for the features obtained using normalized kernels, and this method is called mean-only BN (MOBN) \cite{w_norm}. The results given in Table~\ref{tab:cifar10woDA} show that we can further boost the performance using MOBN for different manifolds. We should also notice that, MOBN boosts the performance only if normalized kernels are used for training, such that the error of SK (MOBN) is 8.52\% while the error of SK (BN) is 8.05\%.

In addition, we compare our methods with the normalized propagation (NormProp) method suggested in \cite{norm_prop}. Briefly, NormProp implements a kernel normalization method using their $\ell_2$ norm at the FC layers, and Frobenius norms at the other convolution layers. In this aspect, they identify kernels as elements of the sphere. However, they do not perform gradient projections and retractions used in SGD steps during BP, but they perform spherical projections during forward propagation. For comparison, we train the same Network in Network (NiN) \cite{nin} architecture utilized in \cite{norm_prop} using our methods. The results show that we obtain similar error for the sphere (9.03\%) compared to their reported error (9.11\%), and we can further boost the performance using the oblique (8.95\%) and the Stiefel manifolds (8.57\%). In addition, proposed methods boost the performance of NiN and SK by 2.29\% and 2.43\%, respectively. Note that, nine convolution layers are used in both NiN and SK, using kernels with different sizes (see \cite{NormProp,w_norm} and sup. mat.). In order to analyze the effect of number of layers to the performance boost, we perform experiments using larger networks in the next section.

\subsection{Results for Training Large-scale CNNs}

We first employ our methods for training of residual networks with constant depth (RCD) and stochastic depth (RSD) consisting of 110 layers \cite{DCCN,SN}. In order to explore how the proposed methods enable us to learn invariance properties as discussed above, we also analyze the results for Cifar and Imagenet datasets that are augmented using standard DA methods (details are given in the sup. mat.). The results given in Table~\ref{tab:rcd}, show that the performance boost is larger for datasets prepared w/o DA compared to the augmented datasets.  In addition, we can further boost the performance even for augmented datasets, since data augmentation is conducted using large scale transformations, while the kernels computed at different layers can learn the invariants at different resolutions. 

\begin{table}[t]
	\centering
	\caption{Results for residual networks (Res) on Cifar-10 with DA.}
	\begin{tabular}{cc}
		\toprule
		\toprule
		
		\textbf{Model} & \textbf{Class. Error (\%)} \\
		\midrule
		\midrule
		
		Res-20 \cite{res_net}  / $\dagger$ & 8.75 / {\color{red} {8.81}} \\		
		Res-20 + MOBN (Sp / Ob / St) $\dagger$ & 8.25 / 8.43 / {\color{blue} \textbf{8.03} } \\
		Res-44 \cite{res_net} / $\dagger$ & 7.17 / {\color{red} {7.16}} \\
		Res-44 + MOBN (Sp / Ob / St) $\dagger$ & 6.99 / 6.89 / {\color{blue} \textbf{6.81}}\\
		\bottomrule
		\bottomrule
	\end{tabular}%
	\label{tab:res10}%
\end{table}%

\begin{table}[t]
	\centering
	\caption{Results for CNNs trained using Imagenet for single crop.}
	\begin{tabular}{C{4.57cm} C{2.750cm}}
		\toprule
		\toprule
		
		\textbf{Model} & \textbf{Top-1 error (\%)} \\
		
		\midrule
		\midrule
		Res-18$\dagger$ / Res-34$\dagger$ / Res-50$\dagger$ & 30.59/26.88/24.52  \\
		PRONG (Inception arch.) \cite{prong} & 28.90 \\
		Res-18+MOBN (Sp / Ob / St)$\dagger$ & 29.13/28.97/{\color{blue}\textbf{28.14}} \\
		Res-34+MOBN (Sp / Ob / St)$\dagger$ & 26.04/25.73/{\color{blue}\textbf{25.16}}\\
		Res-50+MOBN (Sp / Ob / St)$\dagger$  & 23.79/23.70/{\color{blue}\textbf{23.02}}\\
	\bottomrule
	\bottomrule
	\end{tabular}%
	\label{tab:imagenet}%
\end{table}%

Since Res with less number of layers do not perform as well as SK and NiN on Cifar dataset, we also provide the results for the Cifar-10 with DA in Table~\ref{tab:res10}. The results show that our methods can boost the performance of the baseline Res. However, for a smaller Res (Res-20), the kernels of the sphere may outperform the kernels of the oblique as also observed in Table~\ref{tab:rcd}. Moreover, we observe that the amount of performance boost (for St) decreases from 0.78\% to 0.35\%  as the number of layers increases to 44 in Table~\ref{tab:res10}. On the other hand, for St, we obtain 0.65\% and 2.11\% boost for Cifar 10 and 100 with DA, and 0.72\% and 4.98\% boost for Cifar 10 and 100 without DA, using Res consisting of 110 layers equipped with pre-activations (RCD) in Table~\ref{tab:rcd}. Therefore, the amount of boost also depends on the number of classes and use of the augmentation methods. 

We also observe that performance boosts more for Cifar-100 compared to the results obtained for Cifar-10. This result suggests that we can learn feature representations of diverse patterns observed in large number of classes using kernels belonging to the manifolds. In order to scrutinize this observation, we provide the results for training of residual networks (Res) \cite{res_net} using the Imagenet dataset in Table~\ref{tab:imagenet}. The results given in Table~\ref{tab:imagenet} complement the previous observations such that we have 2.45\%, 1.72\% and 1.50\% performance boost (for St) using Res-18, Res-34 and Res-50, respectively. We also provide the performance of a recent method proposed for optimization on a probabilistic manifold of network parameters, called PRONG \cite{prong}. In other words, PRONG implements an approximation for natural gradient descent. An interesting result is that Res-18 with 18 convolution layers, which was trained using our methods with manifolds, outperforms Inception (22 conv. layers) which was trained using PRONG (see Table~\ref{tab:imagenet}).

\newpage
\section{Conclusion}
\label{sec:conc}

We proposed a mathematical framework to explore and make use of geometric properties of spaces of convolution kernels in CNNs. More precisely, we suggested several mathematical methods and tools to describe and utilize kernel spaces by particular topological smooth manifolds, namely embedded and immersed kernel submanifolds. 
Following our theoretical results, we proposed a SGD algorithm for optimization on kernel submanifolds to train CNNs with assurance of convergence to a solution at single minimum of loss.

We employed our algorithm to train the CNNs using benchmark datasets. We observed that our methods boost the performance of the CNNs for various datasets prepared with and without using data augmentation methods. We believe that our results will guide researchers to develop geometry-aware training algorithms that employ powerful regularization methods and take advantage of invariance properties of kernels. In the feature work, we plan to apply our framework for other tasks such as segmentation, detection, pose estimation, action recognition and video analysis. Moreover, we will use our algorithm to train other deep networks such as auto-encoders and recurrent neural networks.

{\small
\bibliographystyle{ieee}
\bibliography{myref}
}

\end{document}